\documentclass[pdflatex,sn-mathphys-num]{sn-jnl}% Math and Physical Sciences Numbered Reference Style 
\usepackage{graphicx}%
\usepackage{multirow}%
\usepackage{amsmath,amssymb,amsfonts}%
\usepackage{amsthm}%
\usepackage{mathrsfs}%
\usepackage[title]{appendix}%
\usepackage{xcolor}%
\usepackage{textcomp}%
\usepackage{manyfoot}%
\usepackage{booktabs}%
\usepackage{algorithm}%
\usepackage{algorithmicx}%
\usepackage{algpseudocode}%
\usepackage{listings}%
\usepackage{caption}
\usepackage{subcaption}
\usepackage{xcolor}
\usepackage{lmodern}
\usepackage{comment}
\usepackage{pifont}

\newcommand{\bx}{\textbf{x}}
\newcommand{\bz}{\textbf{z}}
\newcommand{\bc}{\textbf{c}}
\newcommand{\be}{\textbf{e}}
\newcommand{\bE}{\textbf{E}}

\newcommand{\bR}{\textbf{R}}
\newcommand{\bmu}{\boldsymbol{\mu}}
\newcommand{\blv}{\log\boldsymbol{\sigma^2}}
\newcommand{\bW}{\textbf{W}}

\newcommand{\bZ}{\textbf{Z}}
\newcommand{\btW}{\boldsymbol{\mathcal{W}}}
\newcommand{\btT}{\boldsymbol{\mathcal{T}}}
\newcommand{\btA}{\boldsymbol{\mathcal{A}}}
\newcommand{\btB}{\boldsymbol{\mathcal{B}}}
\newcommand{\btC}{\boldsymbol{\mathcal{C}}}
\newcommand{\Dsyn}{\mathcal{D}_{\text{syn}}}
\newcommand{\Dreal}{\mathcal{D}_{\text{real}}}

\newcommand{\by}{\textbf{y}}
\newcommand{\ttl}{\texttt{TCL}}
\newcommand{\ttt}{\texttt{Transformer}}
\newcommand{\ttd}{\texttt{Linear}}
\newcommand{\tcfd}{$\text{TensorConFormer}^{(\text{Dec})}$}
\newcommand{\tcfe}{$\text{TensorConFormer}^{(\text{Enc})}$}
\newcommand{\cmark}{\ding{51}}
\newcommand{\xmark}{\ding{55}}

\begin{document}

\title[Tabular data generation with tensor contraction layers and transformers]{Tabular data generation with tensor contraction layers and transformers}


\author*[1, 3]{\fnm{Aníbal} \sur{Silva}}\email{up201008538@up.pt}

\author[2, 3]{\fnm{André} \sur{Restivo}}\email{arestivo@fe.up.pt}

\author[2, 3]{\fnm{Moisés} \sur{Santos}}\email{mrsantos@fe.up.pt}
\author[2, 3]{\fnm{Carlos} \sur{Soares}}\email{csoares@fe.up.pt}

\affil[1]{\orgdiv{Faculty of Sciences}, \orgname{University of Porto}, \orgaddress{\street{R. do Campo Alegre}, \city{Porto}, \postcode{4169-007}, \country{Portugal}}}

\affil[2]{\orgdiv{Faculty of Engineering}, \orgname{University of Porto}, \orgaddress{\street{R. Dr. Roberto Frias}, \city{Porto}, \postcode{4200-465}, \country{Portugal}}}

\affil[3]{\orgdiv{LIACC}, \orgname{Artificial Intelligence and Computer Science Laboratory}, \orgaddress{\street{R. Dr. Roberto Frias}, \city{Porto}, \postcode{4200-465}, \country{Portugal}}}

%\affil[2]{\orgdiv{Department}, \orgname{Organization}, \orgaddress{\street{Street}, \city{City}, \postcode{10587}, \state{State}, \country{Country}}}

%\affil[3]{\orgdiv{Department}, \orgname{Organization}, \orgaddress{\street{Street}, \city{City}, \postcode{610101}, \state{State}, \country{Country}}}


%\abstract{The abstract serves both as a general introduction to the topic and as a brief, non-technical summary of the main results and their implications. Authors are advised to check the author instructions for the journal they are submitting to for word limits and if structural elements like subheadings, citations, or equations are permitted.}

% 1. context: data generation is important for multiple purposes (e.g. augmentation, privacy-preserving sharing)
% 2. motivation: VAEs are a common data generation architecture; despite more recent approaches are more popular than VAEs, they have some interesting features (e.g. computational cost) 
% 3. ... recent VAE-based approaches for data generation  replace the linear layers with transformers
% 4. ... motivation: The flexibility of these approaches may be limited by the direct use of transformers, as the architecture is determined by the number of features in the data.
% 5. Our work: To address this limitation, we include a tensor contraction layer, which allows the control of dimensions of the transformer. 
% 6. summary exp setup
% 7. summary of results

\abstract{Generative modeling for tabular data has recently gained significant attention in the Deep Learning domain. Its objective is to estimate the underlying distribution of the data. However, estimating the underlying distribution of tabular data has its unique challenges. Specifically, this data modality is composed of mixed types of features, making it a non-trivial task for a model to learn intra-relationships between them. One approach to address mixture is to embed each feature into a continuous matrix via tokenization,
while a solution to capture intra-relationships between variables is via the transformer architecture. In this work, we empirically investigate the potential of using embedding representations on tabular data generation, utilizing tensor contraction layers and transformers to model the underlying distribution of tabular data within Variational Autoencoders. Specifically, we compare four architectural approaches: a baseline VAE model, two variants that focus on tensor contraction layers and transformers respectively, and a hybrid model that integrates both techniques. Our empirical study, conducted across multiple datasets from the OpenML CC18 suite, 
compares models over density estimation and Machine Learning efficiency metrics. The main takeaway from our results is that leveraging embedding representations with the help of tensor contraction layers improves density estimation metrics, albeit maintaining competitive performance in terms of machine learning efficiency.
}



\maketitle
\section{Introduction}\label{sec:introduction}

Deep Learning has been thoroughly investigated over the last decades and has been successfully applied to various learning tasks. Generative modeling is no exception. Generally, this class of models aims to estimate the underlying distribution of the data. Existing generative flavours encompass variational inference~\cite{Kingma2013AutoEncodingVB, 10.5555/3495724.3496298, pmlr-v37-rezende15}, generative adversarial networks~\cite{NIPS2014_5ca3e9b1} and score-based matching~\cite{10.5555/3454287.3455354}. 

Despite mostly focusing on data modalities such as image~\cite{Bauer2024ComprehensiveEO} and text~\cite{10.1145/3649449}, there has been a recent surge of interest in generative models for tabular data. The interest lies in generating synthetic data to overcome challenges such as data scarcity, missing-value imputation, and individual privacy-preserving (see e.g.,~\cite{r2024navigatingtabulardatasynthesis} for a thorough review).

\paragraph{Challenges in tabular data generation}

Modeling the joint distribution of tabular data has its unique challenges. The main research interest in the generative model is on images, and, usually, the theory behind it assumes a continuous distribution of the data. This is not true for tabular data, which generally presents a mixture of both continuous and discrete variables. Moreover, continuous variables might exhibit several modes and discrete variables may have a considerable number of categories, making it difficult for a given neural network to learn relationships between these two different types of data adequately. This is further supported from a supervised learning perspective, where recent studies showed that tree-based models are still the \textit{de facto} algorithms for classification and regression tasks~\cite{10.5555/3600270.3600307, ye2024closerlookdeeplearning}. 

\paragraph{Tokenization and Tensor Contraction Layers}

A data point of a tabular dataset is usually represented as a heterogeneous vector composed of numerical and discrete features. One possible solution to overcome this heterogeneity is to embed each feature into a matrix via tokenization~\cite{Gorishniy2021RevisitingDL, Zhang2023MixedTypeTD}. In essence, this transformation linearly projects each feature into a continuous vector.  Given this unified continuous matrix representation, one cannot apply a conventional linear transformation to it since weights in a linear layer of a neural network are themselves matrices. Thus, we need to consider high-order mathematical objects to handle embeddings, also known as tensors. To propagate embedding representations along a neural network, we propose using Tensor Contraction Layers~\cite{Kossaifi2017TensorRN}, where weights and biases of this layer are generalized to have arbitrary dimensions. To the best of our knowledge, these types of layers are under-explored in the tabular domain.

\paragraph{Transformers}

The transformer architecture~\cite{10.5555/3295222.3295349} was initially proposed for machine translation and later applied to text generation~\cite{Radford2018ImprovingLU}. Given its unprecedented success, adaptations have been made to this architecture in the past few years for images~\cite{10.1145/3505244}, time series~\cite{10.24963/ijcai.2023/759}, and, naturally, tabular data~\cite{Huang2020TabTransformerTD, Somepalli2021SAINTIN, Zhang2023MixedTypeTD}. In the tabular domain, the transformer architecture purpose is to capture meaningful relations between feature representations of the data via attention mechanisms. 
%To feed tabular data into a transformer, we need to modify its vector representation into a matrix, e.g., via tokenization, where each feature is projected into a vector. The concatenation of these feature representations maps a data point into a continuous embedding matrix.

\paragraph{}

In this work, we empirically investigate the potential of using embedding representations on tabular data generation, leveraging them using tensor contraction layers and transformers. To deal with the heterogeneity of tabular data, tokenization is used, while intra-relationships between feature representations are captured via transformers. %Embedding representations are handled using tensor contraction layers. To the best of our knowledge, these types of layers are under-explored in the tabular data domain. 
For data generation, the Variational Autoencoders (VAEs) architecture is considered. In addition to its base implementation, we also consider — \textit{i)}~TensorContracted, a VAE that encodes tabular data into embeddings and handles them via tensor contraction layers; \textit{ii)}~Transformed, a VAE based on transformers adapted from~\cite{Zhang2023MixedTypeTD}; and \textit{iii)}~TensorConFormer, a VAE that leverages both tensor contraction layers and Transformers. Experiments are conducted using the OpenML CC18 suite~\cite{oml-benchmarking-suites}.

The key findings of this work are: 1)~On average, TensorContracted provided better results than its base implementation over density estimation metrics; 2)~TensorConFormer posits a higher ability to synthesize diverse data; 3)~Transformed fails to generalize the distribution of the data w.r.t. other variations; 4)~Except for Transformed, models still compete regarding Machine Learning efficiency.

%A motivational example, considering 2-dimensional toy datasets with handily crafted decision boundaries is provided in the Appendix \ref{appendix:toydatasets}.

We conclude the introduction by pointing out the contributions of this work. They are

\begin{itemize}
    \item The introduction of Tensor Contraction Layers in tabular data modality as a solution to handle embedding representations and learn more complex relationships;
    \item An empirical study that compares VAE architectures enunciated above based on density estimation and machine learning efficiency metrics.
\end{itemize}

The remainder of this work is structured as follows: In the following Section~\ref{sec:related}, we introduce related works. In Section~\ref{sec:prob_def}, we state the problem definition and the formulation needed to introduce the considered methods, which are detailed in Section~\ref{sec:methods}. The experimental setup is described in Section~\ref{sec:exp_setup}, namely the datasets used throughout this work and the considered evaluation metrics. In Section~\ref{sec:results}, we present our findings. Finally, Section~\ref{sec:conclusions} concludes this work.

\section{Related Work}\label{sec:related}

Lines of work on tabular data generation using Deep Learning methods often involve adapting established architectures from other data modalities.

\paragraph{Generative Adversarial Networks}

A prominent architecture used for this purpose is Generative Adversarial Networks (GANs). In short, GANs consist of a minimax two-player game~\cite{NIPS2014_5ca3e9b1}, where a generator aims at modeling the underlying distribution of the data, while the discriminator discerns if data fed into is real or generated. Given its success in synthesizing high-quality images,  several adaptations to this architecture to handle tabular data were proposed over recent years. One of the earliest examples is medGAN~\cite{Armanious2018MedGANMI}, specifically designed for medical records. Another well-known tabular adaptation of GANs is Conditional GAN (CTGAN)~\cite{10.5555/3454287.3454946}. Here, the authors propose two methods — one to deal with the non-trivial probability density estimation of continuous variables with multiple modes, based on Gaussian Mixture Models, and another to deal with class imbalance present in categorical ones based on a training-by-sampling procedure. Successors to CTGAN include CTAB-GAN~\cite{zhao21} and its enhanced version, CTAB-GAN+~\cite{zhao2023ctab}. CTAB-GAN incorporates additional loss terms to improve the realism of synthetic data, while CTAB-GAN+ refines its predecessor.

\paragraph{Diffusion Models}

Another class of generative models in Deep Learning models is Diffusion Models~\cite{10.5555/3045118.3045358, 10.5555/3495724.3496298}, inspired by principles from non-equilibrium thermodynamics. These models involve two processes --- a forward that gradually adds noise (using a well-defined probability distribution) to data, rendering it indistinguishable from random noise, and a reverse process that gradually recovers data initially sampled from noise. Its first adaptation to the tabular domain was TabDDPM~\cite{10.5555/3618408.3619133}, where the authors propose a mixture of Gaussian and multinomial diffusion to handle numerical and categorical variables, respectively. 
Concurrently, CoDi~\cite{Lee2023CoDiCC} and StaSy were introduced. While the first handles the diffusion process of numerical and categorical variables independently, but conditioned to each other, the latter is based on Score Matching~\cite{10.5555/3454287.3455354} with Self-Paced Learning~\cite{NIPS2014c60d060b}. More recently, a Latent Score-Based Generative Model~\cite{10.5555/3540261.3541124} was successfully applied to tabular data and dubbed TabSyn~\cite{Zhang2023MixedTypeTD}.

\paragraph{Variational Autoencoders}

Finally, another class of generative models that stood its ground over the years in the Deep Learning domain are Variational Autoencoders~\cite{Kingma2013AutoEncodingVB}. This class of models aims to estimate the underlying distribution of the data via variational inference. Adaptations of this model include TVAE~\cite{10.5555/3454287.3454946}, introduced alongside CTGAN, and VAEM~\cite{10.5555/3495724.3496667}, which consists of a two-stage training — the first stage independently trains each feature using a VAE, and the second model's inter-variable dependencies using the latent variables learned from the first stage. Another variation, GOGGLE~\cite{Liu2023GOGGLEGM} was introduced as a generative model that approximates relational structure between variables via Graph Neural Networks, jointly training these relations with a VAE.

\section{Formulation}\label{sec:prob_def}

In the context of tabular data, datasets typically consist of mixed-type variables. In this paper, we focus on datasets that contain numerical and categorical features and those exclusively composed of either type. A formulation of a dataset consisting of a mixture of these features follows.

Let $\mathcal{I} = \{\bx, y\}$ denote an instance of a dataset $\mathcal{D}$ with size $N$. We denote a data point $\bx$ to be represented  as a set of numerical $\bx^{(\text{num})} \in \mathbb{R}^{M_n}$ and categorical features $\bx^{(\text{cat})} \in \mathbb{R}^{M_c}$ as the following vector:

\begin{equation}
    \bx = \left(x_1^{(\text{num})}, ..., x_{M_n}^{(\text{num})}, x_1^{(\text{cat})}, ..., x_{M_c}^{(\text{cat})}\right) \in \mathbb{R}^{M}~~,
\end{equation}

\noindent with $M=M_n+M_c$. Categorical variables $x_j^{(\text{cat})}$ are represented by a one-hot encoded vector, $\bx_j^{(\text{ohe})} \in \mathbb{N}^{|C_j|}$, where $C_j =\{1,..., |C_j|\}$ and $|C_j|$ denotes the number of categories of a given categorical feature $j$ such that in the end, each data point is represented as

\begin{equation}\label{eq:x_final}
    \bx = \left(x_1^{(\text{num})}, ..., x_{M_n}^{(\text{num})}, \bx_1^{(\text{ohe})}, ..., \bx_{M_c}^{(\text{ohe})}\right)~~ \in \mathbb{R}^{M'}~~,
\end{equation}

\noindent where $M' = M_n + \sum_{j=1}^{M_c} |C_j|$.

Finally, throughout this work, we consider conditional data generation. Let $\bc = (c_1, c_2, ..., c_C)$ be a vector of conditional features. Given a generative model $G$, we aim to generate samples from it, conditioned to $\bc$, i.e., $\tilde{\bx} = G(\bx |\bc)$. This is accomplished by training $G$ to estimate the conditional likelihood $p(\bx | \bc)$. In our experiments, we condition the data into the target variable $y$.

%From a modeling perspective, this is done by concatenating a given representation $\boldsymbol{r}$ to $\bc$, that we done by $\boldsymbol{r} \oplus \bc$. 

\subsection{Embeddings Representation}\label{subsec:embvae}

As previously mentioned, one of the challenges in tabular data generation is to properly model its distribution due to the mixed-type nature of features. In this work, we tackle this problem by representing each feature as a continuous vector via tokenization~\cite{Gorishniy2021RevisitingDL, Zhang2023MixedTypeTD}.

\subsubsection{Feature Tokenizer}\label{subsubsec:feature_token}

Let $\bx$ be the input of a neural network. A feature tokenizer takes as input a data point and projects it into a $(M \times d)$-dimensional space as:

\begin{equation}
\begin{split}
    \be_i^{(\text{num})} &= x_i^{(\text{num})} \textbf{w}_i^{(\text{num})} + \textbf{b}_i^{(\text{num})} \\
    \be_i^{(\text{cat})} &= \bx_i^{(\text{ohe})} \textbf{W}_i^{(\text{cat})} + \textbf{b}_i^{(\text{cat})}
\end{split} \label{eq:tokenizer}~~,
\end{equation}

\noindent where $\textbf{w}_i^{(\text{num})}, \textbf{b}_i^{(\text{num})}, \textbf{b}_i^{(\text{cat})} \in \mathbb{R}^{1 \times d}$ and $\textbf{W}_i^{(\text{cat})} \in \mathbb{R}^{|C_j| \times d}$. In other words, each numerical feature is projected into a vector space where each sample shares the same weights, while for categorical features this tokenization acts as a lookup table, i.e., each category has its own set of learnable weights. In the end, $\bx$ is represented as the embedding matrix $\bE \in \mathbb{R}^{M \times d}$ by concatenating each $\be_i$ along the feature dimension, $\bE = \mathop{\bigoplus}_{i=1}^M \be_i$.

Although this type of representation was originally motivated to adapt tabular data into transformers, it may also offer advantages in terms of dimensionality compression. While multiple encoding methods exist for categorical features, one-hot encoding seems to be the most prevalent for generative modeling. However, one-hot encoding high-cardinality features can be cumbersome, as it significantly increases the dimensionality of input data and consequently, the number of parameters in the first layer of a given NN. This tokenization method maintains the feature dimensionality, with an increase of $ d(M' + M)$ parameters into the NN. 
%Also, the additional degree of freedom in the model might be beneficial to learning more complex data representations (although this claim is not explored in this work).

\subsubsection{Feature Detokenizer}\label{subsubsec:feature_detoken}

Given a reconstructed embedding matrix $\Tilde{\bE} \in \mathbb{R}^{M \times d}$, the reconstructed representation $\Tilde{\bx}$ is obtained by projecting each embedding vector $\be_i$ back to the feature space as

\begin{equation}
\begin{split}
    \tilde{x}_i^{(\text{num})} &= \tilde{\be}_i^{(\text{num})} \tilde{\textbf{w}}_i^{(\text{num})} + \tilde{b}_i^{(\text{num})} \\ 
    \tilde{\bx}_i^{(\text{ohe})} &= \text{Softmax}\left(\tilde{\be}_i^{(\text{cat})} \tilde{\bW}_i^{(\text{cat})} + \tilde{\boldsymbol{b}}_i^{(\text{cat})}\right) \\
\end{split}~~,
\end{equation}

\noindent with $\tilde{b}_i^{(\text{num})} \in \mathbb{R}^{1 \times 1}$, $\tilde{\textbf{w}}_i^{(\text{num})} \in \mathbb{R}^{d \times 1},~ \tilde{\boldsymbol{b}}_i^{(\text{cat})} \in \mathbb{R}^{1 \times |C_j|}$ and $\tilde{\bW}_i^{(\text{cat})} \in \mathbb{R}^{d \times |C_j|}$. 
%By the same reasoning, this detokenization procedure adds $M' \times (d+1)$ number of parameters into the NN. 
In the end, we concatenate every reconstructed feature s.t.

\begin{equation}
     \tilde{\bx} = \left(\tilde{x}_1^{(\text{num})}, ..., \tilde{x}_{M_n}^{(\text{num})}, \tilde{\bx}_1^{(\text{ohe})}, ..., \tilde{\bx}_{M_c}^{(\text{ohe})}\right)~~.
\end{equation}

\subsection{Tensor Contraction Layers}\label{subsec:tll}

%As previously stated, our proposed architecture uses Tensor Contraction Layers (TCLs) \cite{Kossaifi2017TensorRN, Kossaifi2017TensorCL}. 
The introduction of Tensor Contraction Layers (TCLs) was initially motivated to replace a fully connected classification head usually designed for image classification in Deep Neural Networks, overcoming the high number of neurons seen after flattening feature maps obtained from convolutional operations, while keeping their multi-linear representations~\cite{Kossaifi2017TensorCL}. In this paper, we use TCLs to handle embedding representations of tabular data. 

\begin{figure}[ht]
    \centering
    \includegraphics[width=\textwidth]{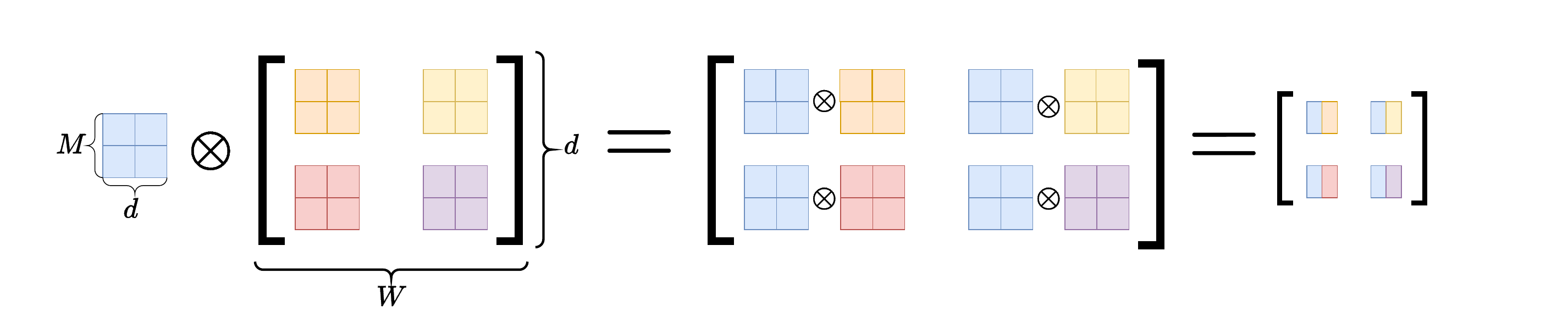}
    \caption{Illustrative example of a tensor contraction operation with $M=d=W=2$.}
    \label{fig:tensor_operation}
\end{figure}

Let a tensor of order $T$ be defined as $\btT \in \mathbb{R}^{t_1 \times t_2 \times ... \times t_T}$. In the tensor formulation, a vector $\textbf{t}$ and a matrix $\textbf{T}$ are a first and second-order tensor, respectively. Next, we define the contraction between two tensors\footnote{To the readers familiar with Einstein summation convention, or numpy Python's package \texttt{einsum} operation, a tensor contraction is defined in our work as \texttt{einsum(`ij, ijkl -> kl',$\bE$, $\btW$)}.}. Let $\btA \in \mathbb{R}^{(a_1 \times ... \times a_A) \times (P_1 \times ... \times P_C)}$ and $\btB \in \mathbb{R}^{(P_1 \times ... \times P_C) \times (b_1 \times ... \times b_B)}$, the contraction between $\btA$ and $\btB$ is defined as

\begin{equation}
    \begin{split}
\btA \otimes \btB &= \sum_{p_1=0}^{P_1}...\sum_{p_C=0}^{P_C} \btA_{a_1, ..., a_A, p_1, ..., p_C} \btB_{p_1, ..., p_C,b_1, ..., b_B} \\
                 &= \btC \in \mathbb{R}^{(a_1 \times ... \times a_A) \times (b_1 \times ... \times b_B)}
    \end{split}~~,
\end{equation}

\noindent given this formulation, in our architectures, weights and biases and fourth and second-order tensors, respectively. A TCL is defined as

\begin{equation}\label{eq:tcl}
    \ttl(\bE) \triangleq \bE \otimes \btW + \textbf{B}~~,
\end{equation}
\\
\noindent with $\bE \in \mathbb{R}^{M \times d}, ~\btW \in \mathbb{R}^{(M \times d) \times (W \times d)}$ and $\textbf{B} \in \mathbb{R}^{W \times d}$~\footnote{Note that in this work we fix the embedding dimension $d$ along tensor contraction operations in a network to be the same as the one used during tokenization.}. An element $w \in W, k \in d$ of the tensor contraction in Eq.~\eqref{eq:tcl}, $o_{wk}$, is calculated as $\sum_i \sum_j \bE_{ij} \btW_{ijwk}$, that outputs a scalar. An example of a tensor contraction operation is illustrated in Fig.~\ref{fig:tensor_operation}, and an analysis of the number of parameters induced by tensor contraction versus linear layers is in Appendix~\ref{appendix:number_of_params}.

\section{Methods}\label{sec:methods}

In this section, we introduce the methods enunciated in Section~\ref{sec:introduction}. We start by motivating and providing a detailed description of TensorConFormer. Following, we present the other variations considered in this work. Table~\ref{tab:layers} summarizes the type of layers present in each architecture. For a review of the theory behind Variational Autoencoders, we recommend the readers to~\cite{10.1561/2200000056}.

\begin{table}[ht]
    \centering
    \caption{Types of layers present in the considered architectures.}
    \begin{tabular}{l|c|c|c}
        \toprule
         \textbf{Model} & Linear & TCL & Transformer \\
        \midrule
        Base & \cmark & \xmark & \xmark \\
        TensorContracted & \xmark & \cmark & \xmark \\
        Transformed & \xmark & \cmark & \cmark \\
        TensorConFormer & \xmark & \cmark & \cmark \\
        \bottomrule
    \end{tabular}
    \label{tab:layers}
\end{table}

\subsection{TensorConFormer}\label{subsec:tensorconformer}

In Section~\ref{sec:prob_def}, we introduced embedding representations as a solution to handle a mixture of types of variables in tabular data. A transformer-based architecture aims to capture relations between variables via attention mechanisms. However, solely relying on this architecture to learn the data distribution might be insufficient as motivated in the toy examples presented in Appendix~\ref{appendix:toydatasets}. Our intuition here is that since the architecture is constrained to the feature dimension of the data, it does not generalize well. To overcome this limitation, we leverage tensor contraction layers to map embeddings into a hidden representation, loosening the constraints imposed by transformers, and allowing the model to learn more fruitful representations by introducing TensorConFormer.

\begin{figure}[!ht]
    \centering
    \includegraphics[width=\textwidth]{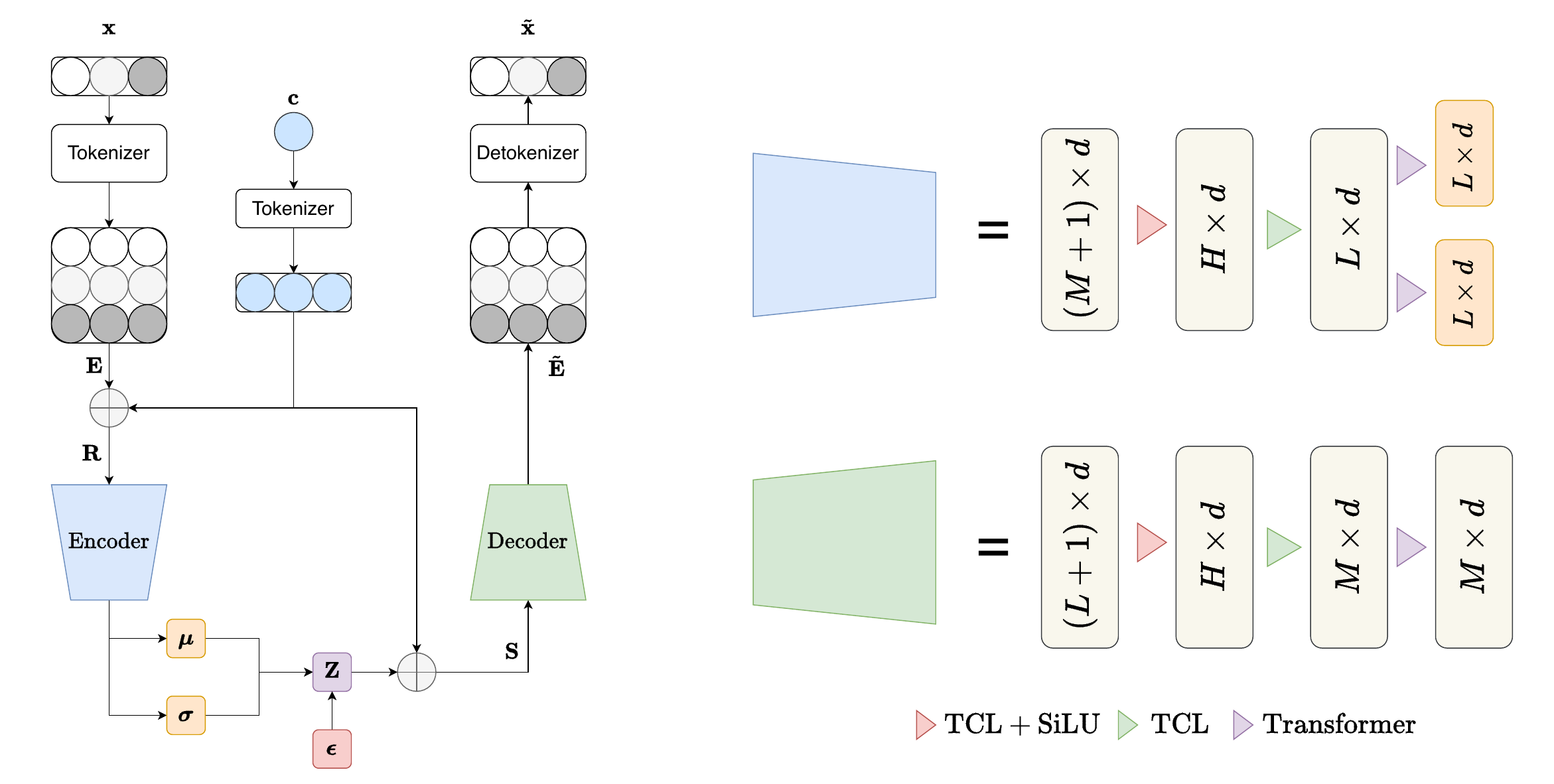}
    \caption{\textbf{Left}: Illustration of an embedding based VAE architecture. \textbf{Right}: Encoder and Decoder mappings of TensorConFormer. Each block denotes a feature map inside an encoder/decoder, with the respective input/output dimensions. Arrows denote operations performed over each feature representation.}
    \label{fig:tensorconformer}
\end{figure}

TensorConFormer consists of transformers that act over the \emph{latent} and \emph{output} space to capture feature relationships via the attention mechanism, while tensor contraction layers are responsible for processing embedding representations guaranteeing an effective solution to deal with high-dimensional representations. The architecture is illustrated in Fig.~\ref{fig:tensorconformer} and a detailed description of a forward pass follows. We begin by projecting $\bR = \bE \oplus \by \in \mathbb{R}^{(M+1) \times d}$ into a hidden representation $\textbf{H} \in \mathbb{R}^{H \times d}$, followed by a projection into the latent representation $\textbf{L}  \in \mathbb{R}^{L \times d}$ via TCLs. Distinct transformers are then used over this representation to learn the statistics of the inference model (or encoder) $q_{\phi}(\bz | \bx, \bc)$. The following set of operations describes the encoder of TensorConFormer

\begin{equation}\label{eq:ttbvae_enc}
    \begin{split}
        \textbf{H} &= \texttt{SiLU}(\ttl(\bR)) \\
        \textbf{L} &= \ttl(\textbf{H}) \\
        (\bmu, \blv) &= \ttt(\textbf{L})
    \end{split}~~,
\end{equation}

\noindent where \texttt{SiLU} is the activation function~\cite{Elfwing2017SigmoidWeightedLU}. Then, we obtain $\bZ$ via the reparametrization trick, $\bZ = \bmu + \boldsymbol{\sigma} \odot \boldsymbol{\epsilon}$, where $\boldsymbol{\epsilon} \sim \mathcal{N}(\boldsymbol{0}, \textbf{I})$. During the generative process (or decoding phase), we aim at learning $p_{\theta}(\bx | \bz, \bc)$. Similarly, we concatenate $\bZ$ with $\by$ along the feature dimension, $\textbf{S} = \bZ \oplus \by \in \mathbb{R}^{(L+1) \times d}$ and project it into the hidden representation $\tilde{\textbf{H}}$, followed by a projection into a representation that shares the same dimensions as $\bE$. The reconstructed embeddings $\Tilde{\bE}$ are finally obtained via a transformer head. The following operations summarize the decoder

\begin{equation}\label{eq:ttbvae_dec}
    \begin{split}
        \Tilde{\textbf{H}} &= \texttt{SiLU}(\ttl(\textbf{S})) \\
        \tilde{\bE} &= \ttt(\ttl(\Tilde{\textbf{H}}))
    \end{split}~~.
\end{equation}

The reconstruction $\Tilde{\bx}$ is obtained via the Detokenization procedure described in Section~\ref{subsubsec:feature_detoken}.

\subsection{Base VAE}\label{subsec:vae}

In the base implementation of VAE, we begin by concatenating $\textbf{r} = \bx \oplus \by \in \mathbb{R}^{M'+N_c}$, where $N_c$ denotes the number of categories in the target variable. The encoder part of the architecture is then followed by

\begin{equation}
    \begin{split}
        \textbf{h} &= \texttt{SiLU}(\ttd(\textbf{r})) \\
        (\bmu, \blv) &= \ttd(\textbf{h})
    \end{split}~~.
\end{equation}

After applying the re-parametrization trick, the decoder projects $\textbf{s} = \textbf{z} \oplus \by \in \mathbb{R}^{L+N_c}$ to the reconstructed output $\tilde{\textbf{x}}$ as $\tilde{\textbf{x}} = \ttd(\texttt{SiLU}(\ttd(\textbf{s})))$.

\subsection{TensorContracted}\label{subsec:tensorcontracted}

The tensor contracted version is essentially the same as its linear counterpart, except that initially, we perform the tokenization described in Section~\ref{subsubsec:feature_token}, and linear layers are replaced by tensor contracted ones, i.e.,

\begin{equation}
    \begin{split}
        \textbf{H} &= \texttt{SiLU}(\ttl(\textbf{R})) \\
        (\bmu, \blv) &= \ttl(\textbf{H})
    \end{split}~~,
\end{equation}

\noindent while the decoder is given by $\tilde{\textbf{E}} = \ttl(\texttt{SiLU}(\ttl(\textbf{S})))$. The reconstructed embedding is finally detokenized as described in Section~\ref{subsubsec:feature_detoken} to obtain $\tilde{\textbf{x}}$.

\subsection{Transformed}\label{subsec:transformed}

The transformed-based VAE is a straightforward adaptation of the VAE proposed in~\cite{Zhang2023MixedTypeTD} for conditional learning, where the posterior distribution parameters are learned directly from the input embedding representation. The encoder is given by 

\begin{equation}
    (\bmu, \blv) = \ttt(\textbf{R})~~.
\end{equation}

Note that in a transformer, the output shares the same dimension as the input, i.e., $\bZ \in \mathbb{R}^{(M+1) \times d}$. Feeding $\textbf{S}$ into the decoder transformer yields $\hat{\bE} = \ttt(\textbf{S}) \in \mathbb{R}^{(M+2) \times d}$. To conform the output representation with $\Tilde{\bE}$ we add the extra layer $\Tilde{\bE} = \ttl(\hat{\bE})$.

To conclude this section, and for the sake of completeness, the model is optimized to maximize the variational lower bound~\cite{NIPS20158d55a249}

\begin{equation}\label{eq:loss}
    \mathcal{L}(\phi, \theta; \bx, \bc) = \mathbb{E}_{q_{\phi}(\bz | \bx, \bc)} \left[\log p_{\theta}(\bx | \bz, \bc)\right] - \mathbb{KL} \left[q_{\phi}(\bz | \bx, \bc) || p(\bz)\right]~~,
\end{equation}

\noindent where $\mathbb{KL}[\cdot || \cdot]$ is the Kullback-Leilbler divergence. The reconstruction loss (first term in Eq. \eqref{eq:loss}) is determined by the squared error or cross-entropy if the feature type is numerical or categorical, respectively.

\section{Experimental Setup}\label{sec:exp_setup}

\subsection{Datasets}\label{subsec:datasets}

We use the OpenML CC18 suite~\cite{oml-benchmarking-suites} as a benchmark to evaluate the methods presented in this paper. It is composed of 72 datasets used for classification tasks. From this benchmark, we select 62 datasets, that encompass samples and feature dimensions in the range between $ N \in [500, 96320]$ and $M \in [4, 256]$, respectively. For all datasets, the train and test splits provided by the OpenML CC18 suite are used, and finally, we extract 15\% of the training set which serves as our validation set.

%\paragraph{Pre-Processing}\label{subsubsec:preproc}  
For all datasets, the following pre-processing is applied: 1) we begin by dropping features that only contain missing values, and numerical or categorical features with 0 variance or only one category, respectively; 2) numerical and categorical columns with missing values are replaced with the mean and mode, respectively; 3) numerical variables are encoded using a quantile transformer with a Gaussian distribution, while categorical variables are encoded using one-hot encoding.

\subsection{Training Details}\label{subsec:train_details}

We use the Adam optimizer, with a weight decay of $0.9$ and a cosine decay learning rate schedule, starting at $1 \times 10^{-3}$ (without warmup). In addition, the following early stopping strategy is applied --- if the loss evaluated from the validation set does not improve at least by $1 \times 10^{-3}$ w.r.t. the loss obtained from the previous iteration, we increment the patience by 1. The training of a given model stops as patience reaches 25. The batch size is determined w.r.t. the validation set following simple rules.

Regarding model hyperparameters, we keep them constant over all datasets and models unless stated otherwise. Each transformer is defined with one head, two layers, a hidden dimension of 128, and without dropout, following~\cite{Zhang2023MixedTypeTD}. By recommendation~\cite{Gorishniy2021RevisitingDL}, we also use its pre-norm variation~\cite{Wang2019LearningDT}. The embedding dimension $d$ is always set to four. Encoders and decoders based on tensor contraction layers have hidden and latent dimensions $H=96$ and $L=32$, respectively. In contrast, the encoder and decoder based on linear layers have a hidden and latent dimension of $H=512$ and $L=256$.

\subsection{Sampling}\label{subsec:sampling}

After training, each model samples over the latent and target variables as $\bz \sim \mathcal{N}(\boldsymbol{0}, \textbf{I})$ and $y \sim \text{Categorical}(p_1, ..., p_{N_c})$, where $p_i$ is the probability of observing the class $i$ in the training set. In our experiments, the number of synthetic samples shares the same size as the training data.

\subsection{Evaluation Metrics}\label{subsec:eval_metrics}

The synthetic data produced by the generative models under study are evaluated using several metrics found in the literature. We divide the considered metrics into two groups: 1)~\textit{Density Estimation and Statistical}, where statistical and density estimation measurements such as marginals distributions are considered and compared between real and synthetic data; 2)~\textit{Machine Learning-Efficiency}, aiming to determine the usefulness of synthetic data in downstream tasks such as classification. Note that all metrics are defined on a domain between [0, 1], where the higher, the better the model performance is.

\subsubsection{Density Estimation and Statistical}

Under this class of metrics, we consider 1-way marginals, pairwise correlations, and high-density estimation metrics.

\paragraph{1-Way Marginals} The first metric measures how similar the (independent) feature distributions between real and synthetic data are. The Kolmogorov-Smirnov statistic~\cite{Hodges1958TheSP} is computed for numerical columns, under the null hypothesis that real and synthetic data are drawn from the same probability distribution, while the Total Variation Distance~\cite{LevinPeresWilmer2006} is applied for categorical ones. In the end, we average the similarities obtained from each feature.

\paragraph{Pairwise-Correlations} Pairwise-correlations measure the dependency between two features in a dataset. Given two columns $(m_1, m_2)$ of both $(\bx, \tilde{\bx})$, if they are both numerical, we determine Pearson's Correlation~\cite{doi:10.1098/rspl.1895.0041}; if they are both categorical, the Contingency Similarity; finally if they are of different types, the numerical column is partitioned into bins, and afterwards, the Contingency Similarity is applied. The score between the correlations $(\rho, \tilde{\rho})$ obtained for each type of data is then determined as

\begin{equation}
    score = 1 - \frac{|\rho_{m_1, m_2} - \tilde{\rho}_{m_1, m_2}|}{2}~~.
\end{equation}

Finally, we average all the scores obtained for each pairwise correlation. For these two first metrics, we use the implementations provided by \texttt{sdmetrics}~\cite{sdmetrics} python's package.

\paragraph{High-Density Estimations} 

These metrics compare the joint distribution of real and synthetic data. We use the work from~\cite{Alaa2021HowFI}, which introduces the notion of $\alpha$-Precision and $\beta$-Recall. Generally speaking, $\alpha$-precision and $\beta$-recall characterize the fidelity and diversity of the generated data w.r.t. to real one, respectively. While $\alpha$-precision is computed by determining the probability that a generated sample resides in the support of the real-data distribution, $\beta$-recall is computed by determining the probability that a real sample resides in the support of the synthetic data distribution. These metrics are evaluated over increasing $(\alpha, \beta)$-support levels of the real and synthetic distributions in the quantile space. Here, we use the implementation of these evaluation metrics provided by the \texttt{synthcity} package~\cite{doi.org/10.48550/arxiv.2301.07573}.

\subsubsection{Machine Learning-Efficiency}

Regarding Machine Learning-Efficiency (ML-Efficiency), we are interested in both utility and fidelity tasks. The classifier taken into consideration is XGBoost~\cite{10.1145/2939672.2939785}. The evaluation procedure is detailed in Appendix~\ref{appendix:ml_eff}. In the end, a test real set is evaluated over two models --- one trained over real data, $\mathcal{M}_{\text{real}}$ and another trained over synthetic data, $\mathcal{M}_{\text{syn}}$. We denote predictions obtained from $\mathcal{M}_{\text{real}}$ and $\mathcal{M}_{\text{syn}}$ as $\hat{y}^{(\text{real})}$, $\hat{y}^{(\text{syn})}$, respectively.

\paragraph{Utility} By utility, we ask how well a model performs when trained over a synthetic dataset $\Dsyn$ and evaluated under a holdout set from the real dataset $\bx^{(\text{test})}$. As such, we adopt the Train on Synthethic, Test on Real (TSTR) (e.g.~\cite{Esteban2017RealvaluedT}) methodology. Here, predictions are evaluated using accuracy.

\paragraph{Fidelity} By fidelity, we ask how similar the predictions ($\hat{y}^{(\text{real})}$, $\hat{y}^{(\text{syn})}$) are. This metric is also measured in terms of accuracy, i.e.

\begin{equation}
    \text{Fidelity} = \frac{1}{\left|\mathcal{D}^{(\text{test})}\right|} \sum_{i=1}^{\left|\mathcal{D}^{(\text{test})}\right|} \boldsymbol{1}\left(\hat{y}_i^{(\text{real})} = \hat{y}_i^{(\text{syn})}\right)~~,
\end{equation}

\noindent where $\boldsymbol{1}(\cdot)$ is the indicator function.

\subsection{Implementation Details}

Models are implemented with Python's programming language using JAX ecosystem~\cite{jax2018github} and trained on a Google Cloud Platform Virtual Machine with 16GB of RAM and an NVIDIA T4. We have released the implementation of the considered models and main experiments, which can be found in GitHub \footnote{\href{https://github.com/anibalsilva1/TabGenwithTCL}{https://github.com/anibalsilva1/TabGenwithTCL}}.

\section{Results}\label{sec:results}

Given the considered models, after training them with datasets from the OpenML CC18 suite, we generate new data according to the procedure described in Section~\ref{subsec:sampling}. Following, we evaluate the generated data under the considered metrics defined in Section~\ref{subsec:eval_metrics}. 

We start by analyzing results based on averages and then by looking at each evaluation measure individually and from a model comparison perspective using the Bayes Sign Test. Finally, we analyze how the dimensions of the considered datasets influence the results obtained via ranking.

\paragraph{General Comparison}

An initial comparison between models based on averaged results is provided and summarised in Table~\ref{tab:main_res}. Results in bold denote best performers, while results marked with $(**)$ denote competitors against the best performer when results are not statistically significantly different according to the Wilcoxon's Sign Rank Test~\cite{c4091bd3-d888-3152-8886-c284bf66a93a}, with a $p$-value of 0.01. Generally speaking, TensorConFormer obtained the highest mean for most of the considered metrics. 
% However, except for 1-Way Marginals and $\beta$-Recall, results are not statistically significant w.r.t. at least one of the other VAE architectures. 
Additionally, although results are not statistically significant compared to the runner-up (except for 1-Way Marginals and $\beta$-Recall), the runner-up varies (either VAE or TensorContracted), which indicates that TensorConFormer is better overall. 
%In terms of diversity and ML-Efficiency, this indicates that, although TensorConFormer can generate diversified data, it does not add any additional information regarding decision boundary estimation for the considered ML model, questioning the quality of its generated data.
In terms of diversity and ML-Efficiency, these results indicate that TensorConFormer can generate data that is both diversified and of high fidelity. These results also show that, even though this is not the goal of the method, it does not enrich the data (i.e. it does not lead to models with higher accuracy than the original data).
%Moreover, while all variations keep their performance approximately on par with each other, Transformed is the worst performer, positing that a Transformed-based VAE architecture is insufficient for modeling the distribution of the data.
One interesting observation is that Transformed is the worst method, indicating that a Transformed-based VAE architecture is not all you need for modeling the distribution of the data.

\begin{table}[ht]
    \centering
    \caption{Mean results averaged over all datasets from the OpenCC18 suite for the considered models. Results in \textbf{bold} correspond to the top performer for a given metric. In contrast, results marked with $(**)$ denote competitors against the top performer such that results are not statistically significant according to Wilcoxon's Sign Rank Test with a $p$-value of 0.01.}
    \begin{tabular}{l|ccccc}
    \toprule
    Metrics & \textbf{VAE} & \textbf{TensorContracted} & \textbf{Transformed} & \textbf{TensorConFormer} \\
    \midrule
    1-Way Marginals ($\uparrow$) & $0.899 \pm 0.06$ & $0.901 \pm 0.06$ & $0.875 \pm 0.06$ & $\mathbf{0.916 \pm 0.05}$ \\
    Pairwise Corr. ($\uparrow$) & $0.930 \pm 0.06$ & ${0.934 \pm 0.06}^{**}$ & $0.896 \pm 0.06$ & $\mathbf{0.938 \pm 0.06}$ \\
    $\alpha$-Precision ($\uparrow$) & $0.726 \pm 0.22$ & $\mathbf{0.766 \pm 0.23}$ & $0.607 \pm 0.29$ & ${0.730 \pm 0.23}^{**}$ \\
    $\beta$-Recall ($\uparrow$) & $0.308 \pm 0.25$ & $0.321 \pm 0.25$ & $0.245 \pm 0.24$ & $\mathbf{0.379 \pm 0.26}$ \\
    Utility ($\uparrow$) & ${0.786 \pm 0.16}^{**}$ & ${0.781 \pm 0.17}^{**}$ & $0.629 \pm 0.23$ & $\mathbf{0.794 \pm 0.16}$ \\
    Fidelity ($\uparrow$) & ${0.763 \pm 0.17}^{**}$ & ${0.756 \pm 0.17}^{**}$ & $0.615 \pm 0.23$ & $\mathbf{0.773 \pm 0.16}$ \\
    \bottomrule
    \end{tabular}
    \label{tab:main_res}
\end{table}

\paragraph{Model Comparisons}

\begin{figure}
    \centering
    \includegraphics[width=\textwidth]{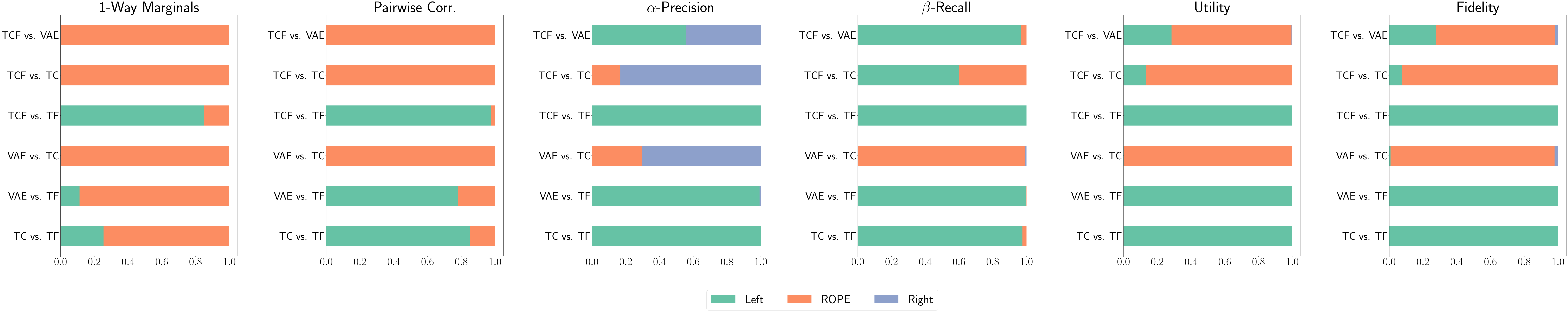}
    \caption{Model comparisons for the considered evaluation metrics using the Bayes Sign Test. Bars denote model comparisons, where each color denotes the probability of a given model (on the left, or right) being practically better than the other, or their performance being practically equivalent using a ROPE of 0.03. TC, TF, and TCF are abbreviations for TensorContracted, Transformed, and TensorConFormer.}
    \label{fig:bst_main}
\end{figure}

Following, we assess the results from a model comparison standpoint using the Bayes Sign Test~\cite{JMLR:v18:16-305}. This test evaluates two models across multiple datasets by assuming that their differences over an evaluation metric are distributed according to a prior probability distribution. Utilizing Bayes Theorem, the test produces a posterior distribution that indicates whether one model is practically superior to the other or equivalent. The latter posterior is called the Region of Practical Equivalence (ROPE), set to 0.03 in our experiments. 
%This means that for a specific dataset, two models are considered equivalent if their performance difference falls within the interval [-0.03, 0.03]. 
The results from this test for the considered evaluation metrics are illustrated in Fig.~\ref{fig:bst_main}.

Starting with 1-Way Marginals, the results indicate that all models are practically equivalent, except that TensorConFormer outperforms Transformed with a probability of approximately 0.85. Despite TensorConFormer showing the highest mean performance (see Table~\ref{tab:main_res}), the Bayesian test doesn't find enough evidence that the difference is significant. For Pairwise Correlations, the base VAE implementation is practically equivalent to both its tensorial counterpart and TensorConFormer. Transformed is the least effective model, as the other variations perform significantly better. Concerning high-order density metrics, the Bayes Sign Test results for $\alpha$-Precision identify TensorContracted as the best performer, being practically superior to VAE, Transformed, and TensorConFormer with probabilities of 0.70, 1, and 0.83, respectively. Conversely, TensorConFormer demonstrates a high probability of being practically better than the other variations in terms of $\beta$-Recall, except when compared to TensorContracted, where their performances are equivalent with a probability of 0.4. For ML-Efficiency, the conclusions for both Utility and Fidelity metrics are similar --- all models, except Transformed, perform equivalently.

\paragraph{Effect of Sample and Feature Size}

\begin{figure}[!ht]
    \centering
    \includegraphics[width=\linewidth]{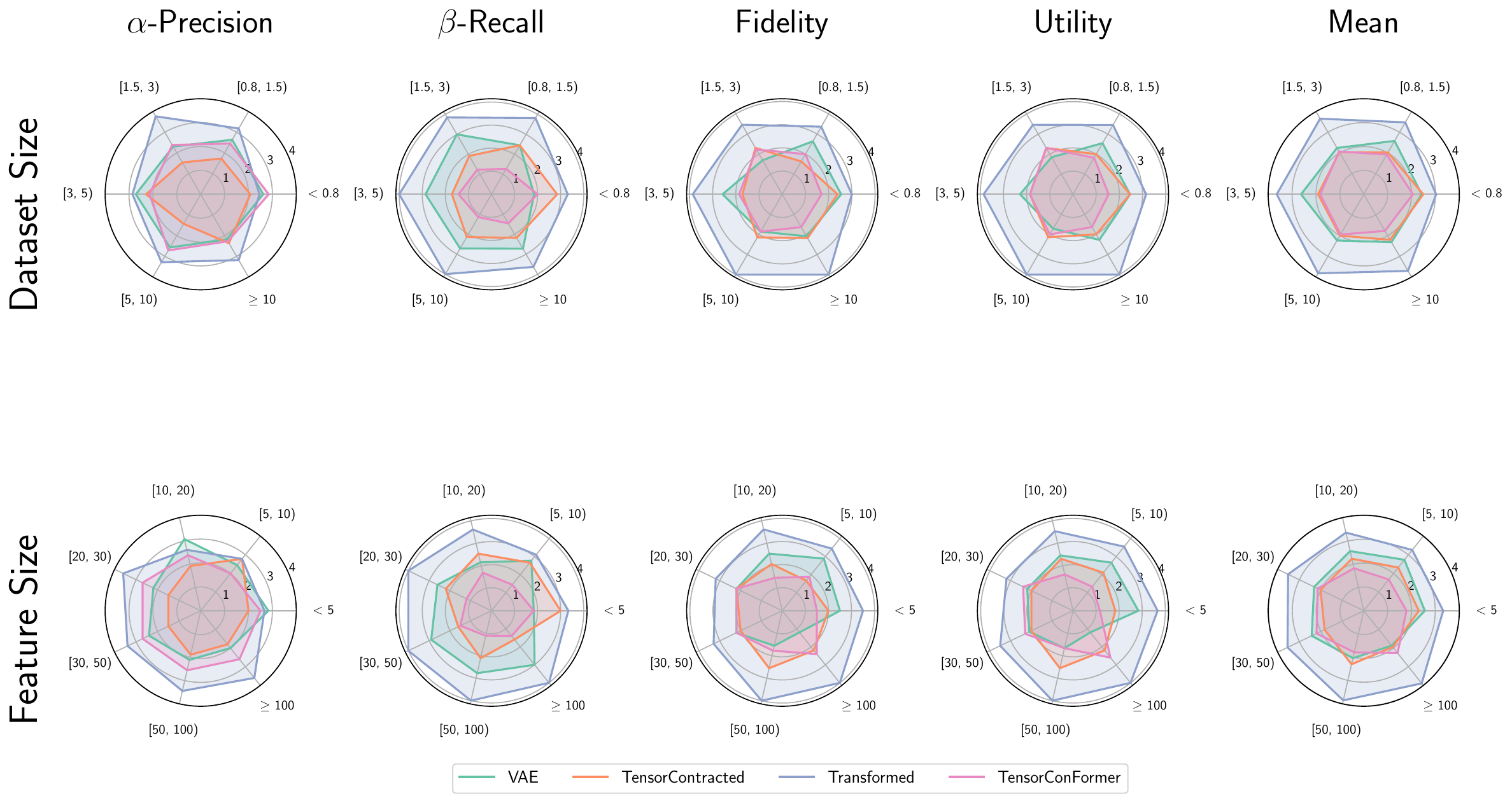}
    \caption{Radar charts for the considered evaluation metrics based on the average ranking of the dataset and feature size (the lower the radius, the better). \textbf{Top}: Average rank as a function of the dataset size (in thousands). \textbf{Bottom}: Average rank as a function of the feature size. The last column denotes the average rank, over all evaluation metrics.}
    \label{fig:avg_rank}
\end{figure}

To conclude our main experiments, we aim to understand whether a model's performance depends on the sample or feature size of a given dataset. Conclusions are drawn by average rank, where we bin datasets into groups based on sample and feature sizes, ensuring approximately equal numbers of datasets per group. We provide a visual representation based on radar charts as depicted in Fig.~\ref{fig:avg_rank}. Note that regarding statistical metrics we only consider high-density metrics for this analysis.

Focusing on $\alpha$-Precision, TensorContracted consistently achieved the lowest average rank, except for datasets of sizes $[3, 5) \times 10^3$ and $ > 10^4$, where it competes with TensorConFormer and the base implementation of VAE. As a function of feature size, TensorContracted also obtained the lowest average rank overall, except over the mid-size range $[5, 10)$, where TensorConFormer was the best performer. Concluding high-density metrics, TensorConFormer consistently obtained the lowest rank for $\beta$-Recall regardless of data or feature size, except for datasets with small feature ($<5$) and sample ($<0.8 \times 10^3$) sizes, where it was surpassed by VAE and had a similar rank. As discussed in Section~\ref{subsec:ablation}, this is posited by the transformer on the decoder part of the network, responsible for modeling intra-relationships between the reconstructed output. Also, as the sample and feature size increase, the base VAE struggles to model the distribution of the data when compared with TensorContracted, hinting that considering embeddings leveraged by TCLs is beneficial in learning a diverse representation of high-dimensional data.

Regarding utility, TensorConFormer obtained the lowest average rank across most dataset size ranges, specifically over low and high dataset sizes. As a function of the feature size, TensorConFormer has the smallest average rank for low-dimensional datasets and competes with both linear and tensor-based architectures for higher dimensional sizes. Focusing on fidelity, the average rank appears to be model-independent, both as a function of dataset and feature sizes.

To conclude this analysis, the last column of Fig.~\ref{fig:avg_rank} presents the average rank obtained for all metrics, calculated by averaging the rank obtained by each model over all metrics, for a given partition. In terms of dataset size, TensorConFormer obtained the lowest average rank, although it competes with TensorContracted in the range of $[0.8, 10)$, while as a function of the feature size, it competes with TensorContracted.

\subsection{Visualization}\label{subsec:vis}

Here, we visually present synthetic data sampled from the considered models. Independent and joint feature distributions are illustrated, and finally, similarities between embedding representations obtained during training are presented.

\paragraph{Feature Distribution}

\begin{figure}[t]
    \centering
    \includegraphics[width=\textwidth]{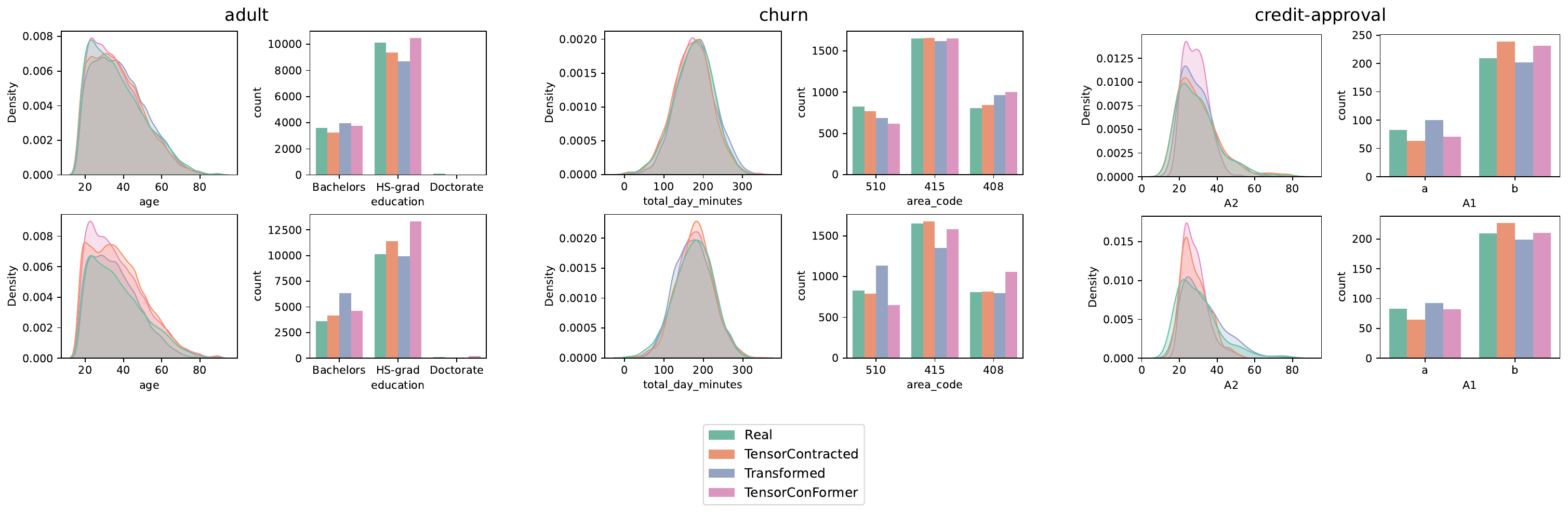}
    \caption{Feature distributions of continuous and categorical variables of pre-selected datasets, conditioned over the majority class. The top row presents the distribution of generated data from the considered models when trained with the whole data, while the bottom row shows the same distribution when the given models are only trained with samples from the majority class.}
    \label{fig:distributions}
\end{figure}

We compare feature distributions of the original data to those of the data generated by the considered models. The similarity of the distributions is an indication of the quality of the generated data. Fig.~\ref{fig:distributions} presents generated feature distributions from the considered models for three datasets from the CC18 suite (\texttt{adult}, \texttt{churn}, and \texttt{credit-approval}). We consider the distribution over the majority class and compare the distributions obtained by training a given model with all data (top-row), and only with data from the majority class (bottom-row). Visually, synthetic data obtained from models trained with all data provided a more similar distribution to real data than models trained only over the majority class.

\paragraph{Data Distribution}

%Next, a visual representation of the data is provided by projecting it into a 2-dimensional space via Uniform Manifold Approximation and Projection for Dimension Reduction (UMAPs) \cite{sainburg2021parametric} (see Fig.\ref{fig:umaps}). 
To compare the generated data with the original one, we first project it into a 2-dimensional space. If the distribution of the generated data in the projected space is similar to that of the original data, it is an indication of the quality of the corresponding generation method.

\begin{figure}[ht]
    \centering
    \includegraphics[width=\linewidth]{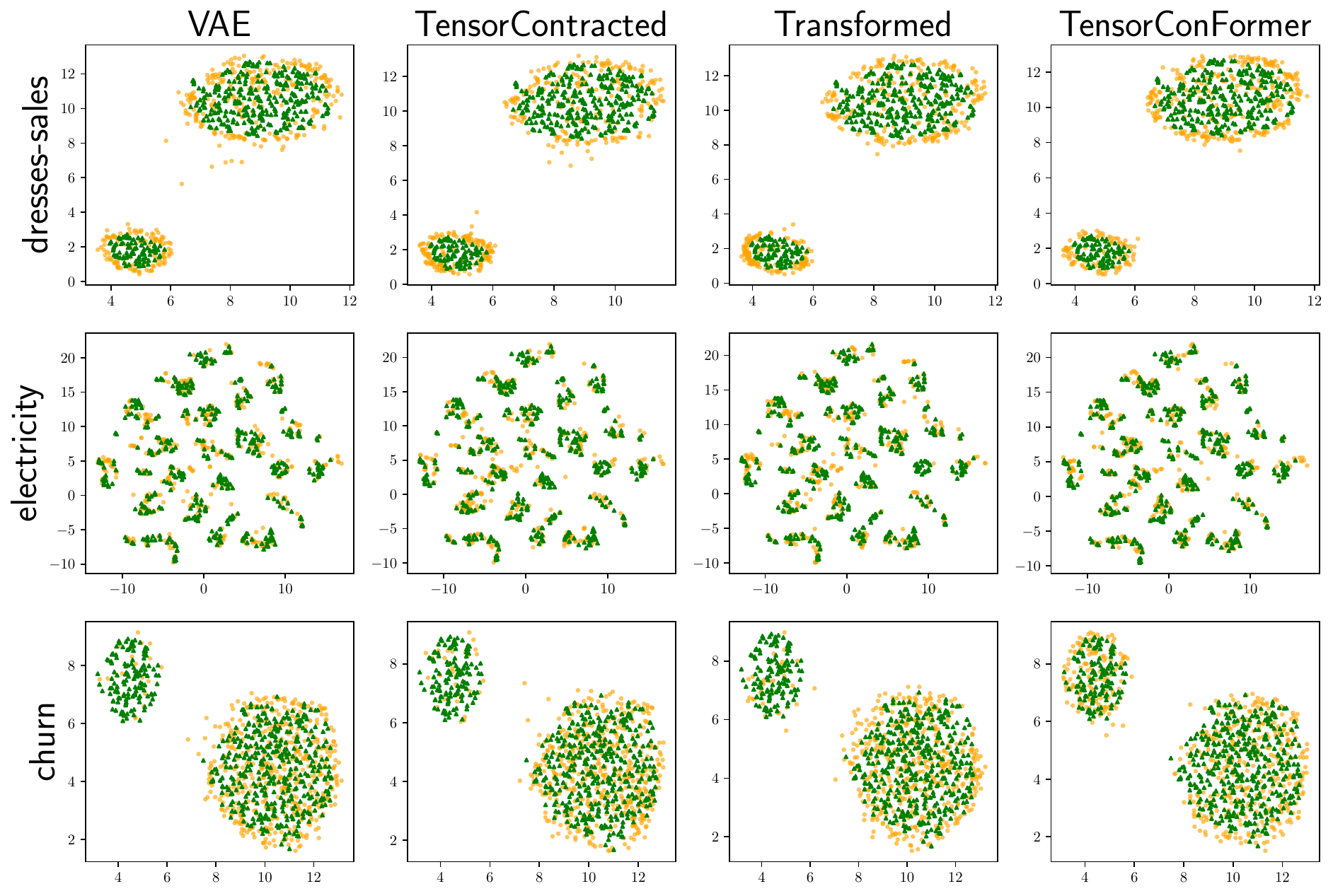}
    \caption{UMAPs for the considered models. Green triangles denote real samples, while orange circles synthetic data.}
    \label{fig:umaps}
\end{figure}

The data was projected using a Uniform Manifold Approximation and Projection for Dimension Reduction (UMAPs)~\cite{sainburg2021parametric} (see Fig.~\ref{fig:umaps}). 
The manifold is estimated using real data, and the synthetic data is then mapped onto it based on this estimation. A visual inspection posits that for both \texttt{dresses-sales} and \texttt{electricity} datasets, the considered models provide similar coverage of the real data. In contrast, for the \texttt{churn} dataset, TensorConFormer had better coverage of real data, specifically over the smallest cluster.

\paragraph{Embeddings Similarities}

Finally, we compare the learned feature representations along training between TensorConFormer and TensorContracted via cosine similarities. Specifically, we look at the output $\Tilde{\textbf{E}}$ and latent $\textbf{Z}$ embedding representations. A detailed description of how these similarities are obtained can be found in Appendix~\ref{appendix:embeddings_similarities}. Fig.~\ref{fig:emb_sims} posits that the output embedding representations tend to be relatively dissimilar at the feature level. Notably, for the \texttt{adult} dataset, feature representation of \texttt{native-country} are highly similar, possibly due to its imbalance over the \texttt{USA} category. Conversely, latent representations are generally identical. This is somewhat expected, as both encoders approximate the respective posterior with $p(\bz)$ (cf. Eq~\eqref{eq:loss}).

\begin{figure}[ht]
    \centering
    \includegraphics[width=\linewidth]{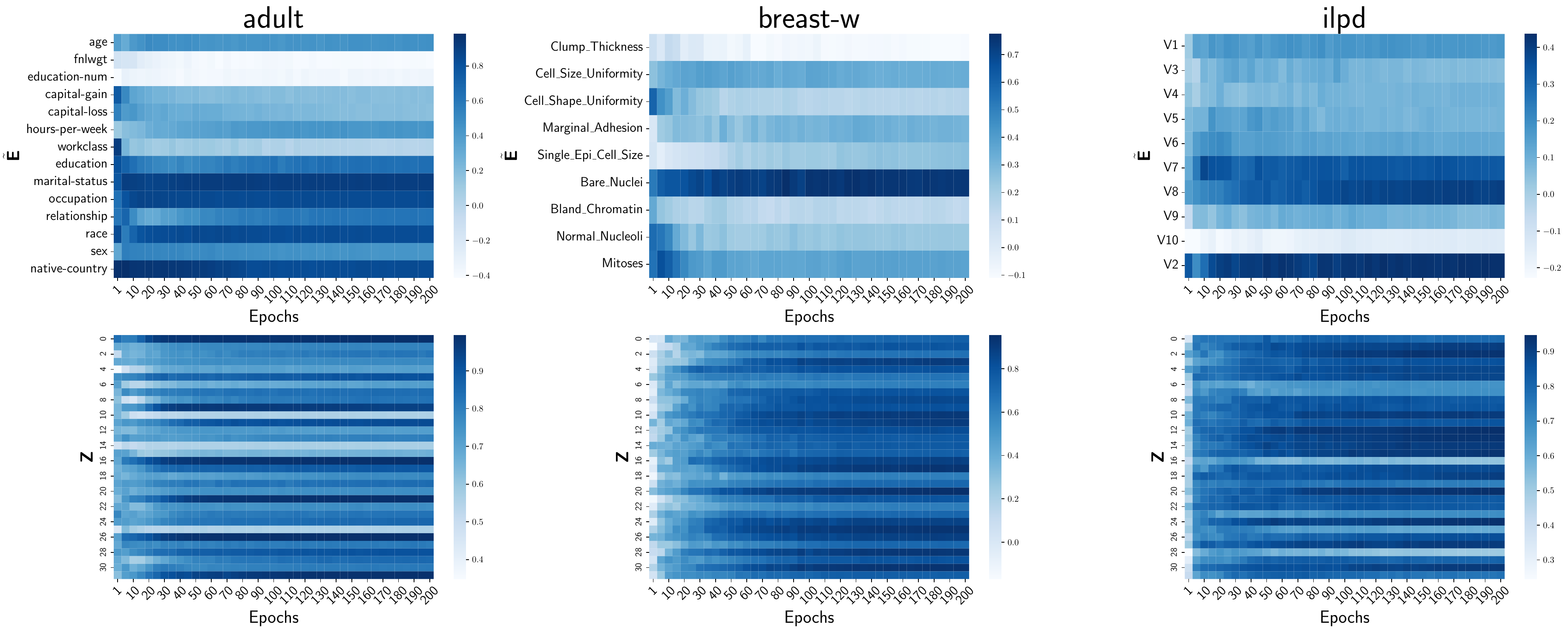}
    \caption{Feature embeddings similarities between TensorContracted and TensorConFormer during training. Each cell represents the cosine similarity of a given embedding representation, averaged over all training samples, for the output reconstructed embedding $\tilde{\textbf{E}}$ (top) and latent space $\textbf{Z}$ (bottom) (cf. Eq. \eqref{eq:avg_sim}).}
    \label{fig:emb_sims}
\end{figure}

\subsection{Ablation Study}\label{subsec:ablation}

To understand the impact of using transformers in the proposed model, we analyze the effect of removing transformers from TensorConFormer encoder and decoder components. Specifically, we consider two ablations. In the first, the parameters of $q_{\phi}(\bz | \bx, \bc)$ are estimated using TCLs, by changing the last line of Eq.~\eqref{eq:ttbvae_enc} into $(\bmu, \blv) = \texttt{TCL}(\textbf{L})$. With the transformer retained in the decoder, we refer to this ablation as $\text{TensorConFormer}^{(\text{Dec})}$. In the second ablation, we remove the transformer head from Eq.~\eqref{eq:ttbvae_dec}, obtaining the reconstructed embedding representation as $\tilde{\bE} = \ttl(\Tilde{\textbf{H}})$. Accordingly, we refer to this ablation as $\text{TensorConFormer}^{(\text{Enc})}$.

\begin{table}[ht]
    \centering
    \caption{Mean results averaged over all datasets from the OpenCC18 suite for the considered ablations. Results in \textbf{bold} correspond to the top performer for a given metric. In contrast, results marked with ($**$) denote competitors against the top performer such that results are not statistically significant according to Wilcoxon's Sign Rank Test with a $p$-value of 0.01.}
    \begin{tabular}{l|ccc}
    \toprule
    Metrics & \textbf{TensorConFormer} & \textbf{\tcfe} & \textbf{\tcfd} \\
    \midrule
    1-Way Marginals ($\uparrow$) & $0.916 \pm 0.05$ & $0.895 \pm 0.06$ & $\mathbf{0.925 \pm 0.04}$ \\
    Pairwise Corr. ($\uparrow$) & ${0.938 \pm 0.06}^{**}$ & $0.928 \pm 0.06$ & $\mathbf{0.941 \pm 0.06}$ \\
    $\alpha$-Precision ($\uparrow$) & ${0.73 \pm 0.23}^{**}$ & ${0.745 \pm 0.24}^{**}$ & $\mathbf{0.756 \pm 0.22}$ \\
    $\beta$-Recall ($\uparrow$) & ${0.379 \pm 0.26}^{**}$ & $0.324 \pm 0.26$ & $\mathbf{0.387 \pm 0.25}$ \\
    Utility ($\uparrow$) & ${0.794 \pm 0.16}^{**}$ & ${0.782 \pm 0.16}^{**}$ & $\mathbf{0.799 \pm 0.15}$ \\
    Fidelity ($\uparrow$) & $\mathbf{0.773 \pm 0.16}$ & ${0.759 \pm 0.17}^{**}$ & ${0.772 \pm 0.16}^{**}$ \\
    \bottomrule
    \end{tabular}
    \label{tab:abla_res}
\end{table}

\begin{figure}
    \centering
    \includegraphics[width=\textwidth]{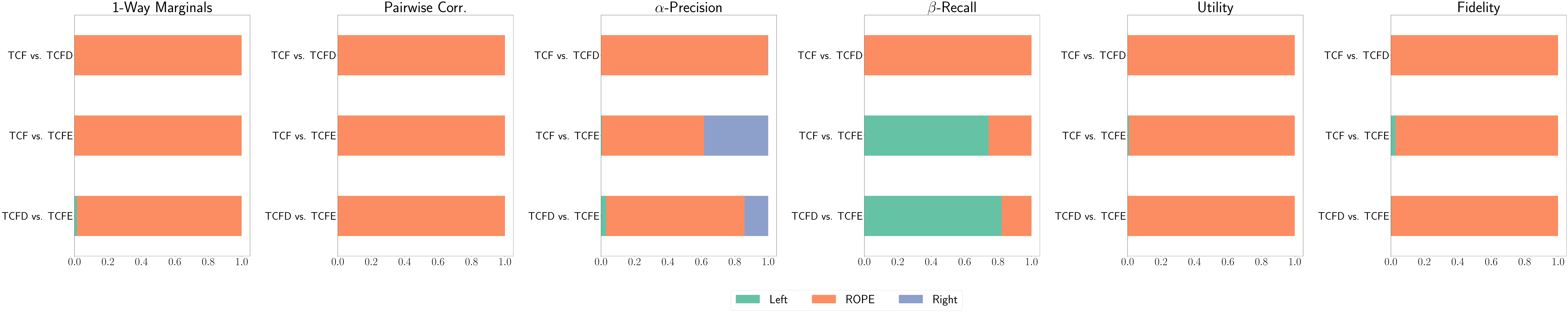}
    \caption{Ablations comparison for the considered evaluation metrics using the Bayes Sign Test. Bars denote a model comparison, where each color denotes the probability of a given model (on the left, or right) being practically better than the other, or their performance being practically equivalent using a ROPE of 0.03. TCF, TCFD, and TCFE are abbreviations for TensorConFormer, \tcfd, and \tcfe, respectively.}
    \label{fig:bst_abla}
\end{figure}

Similarly to the previous study, we begin by drawing conclusions based on the average obtained by a given model over the considered metrics (cf. Table \ref{tab:abla_res}). Results show a performance deterioration when the transformer head is removed from TensorConFormer. In fact, the top performer over most of the considered metrics was \tcfd~(although in most cases not statistically significant w.r.t. TensorConFormer), questioning the necessity of considering transformers to model the posterior parameters.

Finally, conclusions are also drawn based on the Bayes Sign Test. Under this test, all models perform equivalently over the considered metrics, except for high-density estimations as shown in Fig.~\ref{fig:bst_abla}. Specifically, for $\alpha$-Precision, \tcfe~ is practically better than its base model and the other ablation with a probability of $\sim 0.4$ and $0.2$, respectively. Conversely, in terms of diversity, other variations are practically better with a probability of $\sim$ 0.7 and $\sim$ 0.8, respectively, leading to the conclusion that, retaining transformers in the encoder allow the model to capture a more faithful representation of the data, while removing it and leveraging the output representation with it leads to a diverse representation of the data.

\section{Conclusions}\label{sec:conclusions}

In this work, we explored the use of tensor contraction layers and transformers for handling embedding representations in tabular data generation, addressing the inherent challenges of its mixed structure and intra-variable relationships. Three variations of Variational Autoencoders were considered in addition to its linear-based architecture. Based on our experiments, we found that combining tensor contraction layers with transformers enhances the diversity of the generated data, however, when it comes to machine learning utility, the performance remains comparable to other architectural variations. In addition, a VAE architecture that solely relies on transformers does not generalize well the distribution of tabular data.

\begin{table}[!ht]
    \centering
    \caption{Number of parameters, training, and sampling times for the considered models. Results are averaged over all considered datasets.}
    \begin{tabular}{l c c c}
        \toprule
        \textbf{Model} & \textbf{Parameters} (\#) & \textbf{Training Time} (min.) & \textbf{Sampling Time} (sec.) \\
        \midrule
        VAE & $\sim$ 462K & 0.71 & 1.66 \\[3pt]
        Transformed & $\sim$ 81K & 6.03 & 4.18 \\[3pt]
        TensorConFormer & $\sim$ 236K & 4.01 & 3.15 \\[3pt]
        TensorContracted & $\sim$ 278K & 3.99 & 2.34 \\[3pt]
        \midrule
        \tcfe & $\sim$ 233K & 4.95 & 2.42 \\[3pt]
        \tcfd & $\sim$ 280K & 4.52 & 4.18 \\[3pt]
        \bottomrule
    \end{tabular}
    \label{tab:stats}
\end{table}

We also analyze the considered models regarding the number of parameters, training, and sampling times (see Table~\ref{tab:stats}). A comparison between transformer-based models shows that while the number of parameters significantly increases (TensorConFormer $\sim 3 ~\times$ Transformed), time-wise there's a relatively low overhead of $\sim 2~$ min on average for training. Evaluation-wise, TensorConFormer surpassed Transformed over the considered metrics. Comparing the linear and tensor variation of VAE, we reduced the parameters by $\sim 50\%$.

We conclude this paper by presenting some limitations and possible future work directions. We begin by noting that, although we consider embeddings for handling the inherent mixed nature of tabular data inside the model, the reconstruction term of the loss function in Eq.~\eqref{eq:loss} is still calculated as a function of the feature type (i.e. if it is continuous or categorical). A possible working direction is to consider pre-trained embeddings (e.g., obtained by training a self-supervised model), that will serve as our training data to the generative model. Another future working direction relates to leveraging embeddings to learn relationships between feature representations other than the ones provided by attention mechanisms, with the help of TCLs.

\bmhead{Acknowledgements}
This work was partially funded by project AISym4Med (101095387) supported by Horizon Europe Cluster 1: Health, ConnectedHealth (n.o 46858), supported by Competitiveness and Internationalisation Operational Programme (POCI) and Lisbon Regional Operational Programme (LISBOA 2020), under the PORTUGAL 2020 Partnership Agreement, through the European Regional Development Fund (ERDF), Agenda “Center for Responsible AI”, nr. C645008882-00000055, investment project nr. 62, financed by the Recovery and Resilience Plan (PRR) and by European Union -  NextGeneration EU, and FCT plurianual funding for 2020-2023 of LIACC (UIDB/00027/2020 UIDP/00027/2020). The computational resources of Google Cloud Platform were provided by the project CPCA-IAC/AF/594904/2023.

\newpage

\begin{appendices}

\section{Toydatasets}\label{appendix:toydatasets}

\begin{figure}[!ht]
    \centering
    \includegraphics[width=\textwidth]{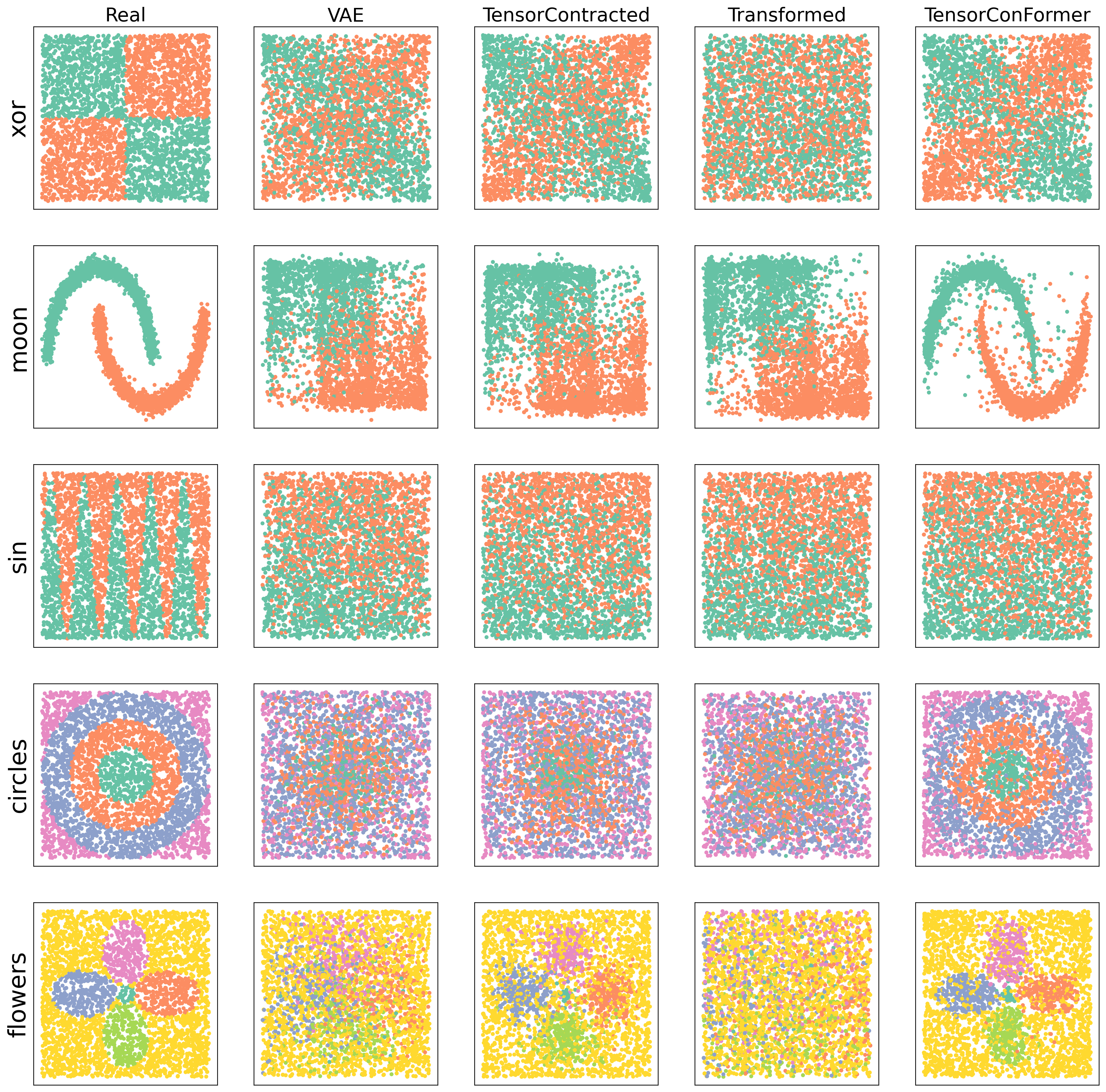}
    \caption{2-Dimensional datasets (first column) with different decision boundaries and the respective generated data from the considered VAE architectures (second to last columns).}
    \label{fig:toydatasets}
\end{figure}

As motivation, we consider 2-dimensional toy datasets with handily crafted decision boundaries. Each dataset considered, depicted in Fig.~\ref{fig:toydatasets}, is constructed by generating $N=20000$ data points uniformly distribution over a given domain, $x_i \sim \mathcal{U}(x^{(i)}_{\min}, x^{(i)}_{\max}),~i={1, 2}$ and then by producing decision boundaries accordingly. For example, in the \texttt{circles} dataset, equations that produce the decision boundaries are

\begin{equation}
    \text{class} = 
    \begin{cases}
        0 & \text{if} ~~ x^2 + y^2 < r^2 \\
        1 & \text{if} ~~ r^2 \leq x^2 + y^2 \leq (2r)^2 \\
        2 & \text{if} ~~ (2r)^2 \leq x^2 + y^2 \leq (3r)^2 \\
        3 & \text{if} ~~ x^2 + y^2 > (3r)^2 \\
    \end{cases}~~.
\end{equation}

All models are trained over 500 epochs, without early stopping.

A comparison between the synthetic data produced for the considered models is depicted in Fig.~\ref{fig:toydatasets}. Overall, TensorConFormer provided a more appropriate conditional distribution of the data, positing the advantage of combining transformers and tensor contraction layers to model complex relationships between features. Interestingly, all the models struggled to sample from the model trained over the \texttt{sin} function. 

\section{Number of Parameters}\label{appendix:number_of_params}

Here, we compare the number of parameters induced by tensor contraction and linear layers (excluding biases terms) when the first linearity is applied to $\bx$ as defined in Eq.~\eqref{eq:x_final}. Applying the tokenization described in Eq.~\eqref{eq:tokenizer} followed by a TCL yields $dM' + MH'd^2$ parameters. On the other hand, for linear layers, we have $M'H$ parameters. The number of parameters between tokenization followed by a TCL and a linear layer applied directly to $\bx$ is equal if its hidden dimension $H$ is given by

\begin{equation}\label{eq:hidden}
    H = d\left(1 + \frac{MH'd}{M'}\right)~~,
\end{equation}

\noindent that scales with $M/M'$. In the limit where $\bx$ is considerably sparse and the hidden dimension of the TCL $H'$ is sufficiently high such that, $MH'd / M' \sim 1$, we have $H \sim 2d$. In the scenario where $M=M'$, $H = d(1+H'd)$. In other words, under a sparse dataset the hidden dimension of a linear layer scales with the embedding dimension $d$, while when the tabular data consists only of numerical features, the hidden dimension of the linear layer scales with the product between the hidden dimension of the tensor contraction layer with the square of the embedding dimension.

As a particular case, we consider the \texttt{adult} dataset from OpenML CC18, with $M=15$ and $M'=107$. In our experiments, we set $H=512$, $H' = 96$ and $d=4$ (cf. Section~\ref{sec:exp_setup}). Using Eq.~\eqref{eq:hidden} we obtain a smaller $H \sim 219$. We show in Section~\ref{sec:results} that the TensorContracted architecture can obtain a superior, or keep performance on par in terms of density estimation metrics w.r.t. its base implementation while reducing the number of parameters $\sim 50\%$ (cf. Table~\ref{tab:stats}).

\section{XGBoost Fine-tuning}\label{appendix:ml_eff}

The fine-tuning procedure utilized for each dataset consists of the following steps: 1) we begin by performing Grid-Search over a given hyper-parameter search space described in Table~\ref{tab:ml_eff_hp}, which is evaluated using 5-fold stratified cross-validation. The accuracy is used to determine the best parameters; 2) Given the best hyper-parameter combination, we re-train the model with the best parameters and evaluate the results in terms of utility and fidelity. These results are always evaluated over a holdout set $\mathcal{D}^{(\text{test})}$, which was not used to train a generative model nor the considered ML model. In addition, when this procedure is being performed under $\Dreal$, these steps are always performed using the same training set used to train a given generative model. 

\begin{table}[ht]
    \centering
    \begin{tabular}{c|c}
         \textbf{Parameter Name} & \textbf{Search Space} \\
         \toprule
         \texttt{n\_estimators} &  [100, 200]  \\
         \texttt{subsample} & [0.7, 0.9, 1] \\
         \texttt{colsample\_bytree} & [0.7, 0.9, 1] \\
         \bottomrule
    \end{tabular}
    \caption{Hyper-parameter space used for XGBoost.}
    \label{tab:ml_eff_hp}
\end{table}

\section{Embeddings similarities}\label{appendix:embeddings_similarities}

The cosine similarity between two vectors $\textbf{a}, \textbf{b}$ is given by

\begin{equation}
    \text{similarity}(\textbf{a}, \textbf{b}) = \frac{\textbf{a} \cdot \textbf{b}}{||\textbf{a}|| \cdot ||\textbf{b}||}~~.
\end{equation}

Defining $\textbf{A} = (\textbf{a}_1, ..., \textbf{a}_S)$, $\textbf{B} = (\textbf{b}_1, ..., \textbf{b}_S)$ as the matrices composed of vectors $\textbf{a}_s, \textbf{b}_s$, the similarity vector between them is given by

\begin{equation}
    \textbf{\text{similarity}}(\textbf{A}, \textbf{B}) = (\text{similarity}(\textbf{a}_1, \textbf{b}_1), ..., \text{similarity}(\textbf{a}_S, \textbf{b}_S))~~.
\end{equation}

If we have $N$ samples for each similarity vector, their average is given by

\begin{equation}\label{eq:avg_sim}
    \frac{1}{N} \sum_{i=1}^N\textbf{\text{similarity}}(\textbf{A}^{(i)}, \textbf{B}^{(i)})~~.
\end{equation}

The similarities presented in Fig.~\ref{fig:emb_sims} are then determined via Eq.~\eqref{eq:avg_sim}, where $\textbf{A}, \textbf{B}$ denotes embeddings representations of two different models (TensorConFormer and TensorContracted), with $\textbf{A}, \textbf{B} = \{\Tilde{\textbf{E}}, \textbf{Z}\}$. 

\section{Dataset Description}

\begin{sidewaystable}
\begin{minipage}{0.5\textwidth}
\centering
\begin{tabular}{ccccc}
\toprule
\textbf{Id} & \textbf{\# Samples} & \textbf{\# Num.} & \textbf{\# Cat.} & \textbf{Task} \\
\midrule
3 & 3196 & 0 & 36 & binary \\
6 & 20000 & 16 & 0 & multi-class \\
11 & 625 & 4 & 0 & multi-class \\
12 & 2000 & 216 & 0 & multi-class \\
14 & 2000 & 76 & 0 & multi-class \\
15 & 699 & 9 & 0 & binary \\
16 & 2000 & 64 & 0 & multi-class \\
18 & 2000 & 6 & 0 & multi-class \\
22 & 2000 & 47 & 0 & multi-class \\
23 & 1473 & 2 & 7 & multi-class \\
28 & 5620 & 64 & 0 & multi-class \\
29 & 690 & 6 & 9 & binary \\
31 & 1000 & 7 & 13 & binary \\
32 & 10992 & 16 & 0 & multi-class \\
37 & 768 & 8 & 0 & binary \\
43 & 4601 & 57 & 0 & binary \\
45 & 3190 & 0 & 60 & multi-class \\
49 & 958 & 0 & 9 & binary \\
53 & 846 & 18 & 0 & multi-class \\
219 & 45312 & 7 & 1 & binary \\
2074 & 6430 & 36 & 0 & multi-class \\
2079 & 736 & 14 & 5 & multi-class \\
3021 & 3772 & 7 & 22 & binary \\
3022 & 990 & 10 & 2 & multi-class \\
3549 & 841 & 70 & 0 & multi-class \\
3560 & 797 & 0 & 4 & multi-class \\
3902 & 1458 & 37 & 0 & binary \\
3903 & 1563 & 37 & 0 & binary \\
3904 & 10885 & 21 & 0 & binary \\
3913 & 522 & 21 & 0 & binary \\
3917 & 2109 & 21 & 0 & binary \\
\bottomrule
\end{tabular}
\end{minipage}%
\hspace{0.5cm}
\begin{minipage}{0.5\textwidth}
\centering
\begin{tabular}{ccccc}
\toprule
\textbf{Id} & \textbf{\# Samples} & \textbf{\# Num.} & \textbf{\# Cat.} & \textbf{Task} \\
\midrule
3918 & 1109 & 21 & 0 & binary \\
7592 & 48842 & 6 & 8 & binary \\
9946 & 569 & 30 & 0 & binary \\
9952 & 5404 & 5 & 0 & binary \\
9957 & 1055 & 41 & 0 & binary \\
9960 & 5456 & 24 & 0 & multi-class \\
9964 & 1593 & 256 & 0 & multi-class \\
9971 & 583 & 9 & 1 & binary \\
9977 & 34465 & 89 & 29 & binary \\
9978 & 2534 & 72 & 0 & binary \\
9985 & 6118 & 51 & 0 & multi-class \\
10093 & 1372 & 4 & 0 & binary \\
10101 & 748 & 4 & 0 & binary \\
14952 & 11055 & 0 & 30 & binary \\
14954 & 540 & 18 & 19 & binary \\
14965 & 45211 & 7 & 9 & binary \\
14969 & 9873 & 32 & 0 & multi-class \\
125920 & 500 & 1 & 11 & binary \\
125922 & 5500 & 40 & 0 & multi-class \\
146195 & 67557 & 0 & 42 & multi-class \\
146800 & 1080 & 77 & 0 & multi-class \\
146817 & 1941 & 27 & 0 & multi-class \\
146819 & 540 & 18 & 0 & binary \\
146820 & 4839 & 5 & 0 & binary \\
146821 & 1728 & 0 & 6 & multi-class \\
146822 & 2310 & 16 & 0 & multi-class \\
146824 & 2000 & 240 & 0 & multi-class \\
167119 & 44819 & 6 & 0 & multi-class \\
167120 & 96320 & 21 & 0 & binary \\
167140 & 3186 & 0 & 180 & multi-class \\
167141 & 5000 & 16 & 4 & binary \\
\bottomrule
\end{tabular}
\end{minipage}
\caption{OpenML CC18 suite datasets description.}
\label{tab:cc18data}
\end{sidewaystable}
\end{appendices}

\newpage


\bibliography{bibliography}% common bib file

\end{document}